\documentclass[runningheads]{llncs}
\usepackage{graphicx}
\usepackage{multirow}
\usepackage{array}

\newcolumntype{P}[1]{>{\centering\arraybackslash}p{#1}}
\begin{document}
\title{Brain Diffuser: An End-to-End Brain Image to Brain Network Pipeline}
%
%


\author{Xuhang Chen\inst{1,3}\and
Baiying Lei\inst{2}\and
Chi-Man Pun\inst{1}\and
Shuqiang Wang\inst{3}
}

\institute{University of Macau \and
Shenzhen University \and
Shenzhen Institutes of Advanced Technology, Chinese Academy of Sciences 
\email{\{yc17491,cmpun\}@umac.mo}\\
\email{leiby@szu.edu.cn}\\
\email{sq.wang@siat.ac.cn}
}

\maketitle              

\begin{abstract}
Brain network analysis is essential for diagnosing and intervention for Alzheimer's disease (AD). However, previous research relied primarily on specific time-consuming and subjective toolkits. Only few tools can obtain the structural brain networks from brain diffusion tensor images (DTI). In this paper, we propose a diffusion based end-to-end brain network generative model Brain Diffuser that directly shapes the structural brain networks from DTI. Compared to existing toolkits, Brain Diffuser exploits more structural connectivity features and disease-related information by analyzing disparities in structural brain networks across subjects. For the case of Alzheimer's disease, the proposed model performs better than the results from existing toolkits on the Alzheimer’s Disease Neuroimaging Initiative (ADNI) database.
\end{abstract}

\section{Introduction}

Medical image computation is an ubiquitous method to detect and diagnose neurodegenerative disease. It is often related with computer vision to aid in the diagnosis of this form of brain disease~\cite{wang2019ensemble,hu2020medical}, using CT or MRI to study the morphological feature. Deep learning has been extensively used in medical image processing. Due to its superior performance, several complicated medical imaging patterns may be discovered, allowing for the secondary diagnosis of illness~\cite{wang2020ensemble}.

Nonetheless, high-dimensional medical image patterns are challenging to discover and evaluate. Brain is a complex network, and cognition need precise coordination across regions-of-interest (ROIs). Hence, brain network is introduced to overcome the above problems. A brain network can be described as a collection of vertices and edges. ROIs of the brain are represented by the vertices of a brain network. The edges of a brain network indicate the interaction between brain ROIs. Brain network as a pattern of interconnected networks may more accurately and efficiently represent latent brain information. There are two fundamental forms of connection in brain networks: functional connectivity (FC) and structural connectivity (SC). FC is the dependency between the blood oxygen level-dependent BOLD signals of two ROIs. SC is the strength of neural fiber connections between ROIs. Many studies have indicated that by employing FC or SC to collect AD-related features, more information may be obtained than conventional neuroimaging techniques~\cite{wang2018classification,jeon2020enriched,wang2017automatic}.

Nevertheless, the aforementioned brain networks are produced by the use of particular preprocessing toolkits, such as PANDA~\cite{cui2013panda}. These toolkits are sophisticated and need a large amount of manual processing steps to go through the pipeline, which might lead to inconsistent findings if researchers make parametric errors or omit critical processing stages. Therefore, we construct an end-to-end brain network generation network to simplify the uniform generation of brain networks and accurate AD development prediction. We summarizes our contribution as follows:
\begin{enumerate}
    \item A diffusion-based model, Brain Diffuser, is proposed for brain network generation. Compared to existing toolkits, it is an end-to-end image-to-graph model that can investigate potential biomarkers in view of brain network.
    \item A novel diffusive generative loss is designed to characterize the differences between the generated and template-based brain networks, therefore minimizing the difference in intrinsic topological distribution. It can considerably increase the proposed model's accuracy and resilience.
    \item To generate brain networks embedded with disease-related information, the latent cross-attention mechanism is infused in Brain Diffuser so that the model focuses on learning brain network topology and features.
\end{enumerate}

\section{Related work}
Recent years have seen the rise of graph-structured learning, which covers the whole topology of brain networks and has considerable benefits in expressing the high-order interactions between ROIs. Graph convolutional networks (GCN) are widely used in graph-structured data such as brain networks. Abnormal brain connections have been effectively analyzed using GCN approaches~\cite{parisot2018disease,ktena2017distance}.
Pan \emph{et al.} introduced a novel decoupling generative adversarial network (DecGAN) to detect abnormal neural circuits for AD~\cite{pan2021decgan}. CT-GAN~\cite{pan2022cross} is a cross-modal transformer generative adversarial network to combine functional information from resting-state functional magnetic resonance imaging~(fMRI) with structural information from DTI. Kong \emph{et al.} introduced a structural brain-network generative model (SBGM) based on adversarial learning to generate the structural connections from neuroimaging~\cite{kong2022adversarial}.

Diffusion Model~\cite{sohl2015deep} is a probabilistic generative models designed to learn a data distribution by gradually denoising a normally distributed variable, which corresponds to learning the reverse process of a Markov Chain. Diffusion models have been used in medical imaging for anomaly detection~\cite{Wolleb2022DiffusionMF,Pinaya2022FastUB} and reconstruction~\cite{Chung2021ScorebasedDM}. And its strong capability in image synthesis also can be used to generated graph data~\cite{jo2022score}. Therefore we adopt diffusion model as our generation module to synthesize brain networks.

\begin{figure*}[ht]
    \begin{minipage}[b]{1.0\linewidth}
        \includegraphics[width=\linewidth]{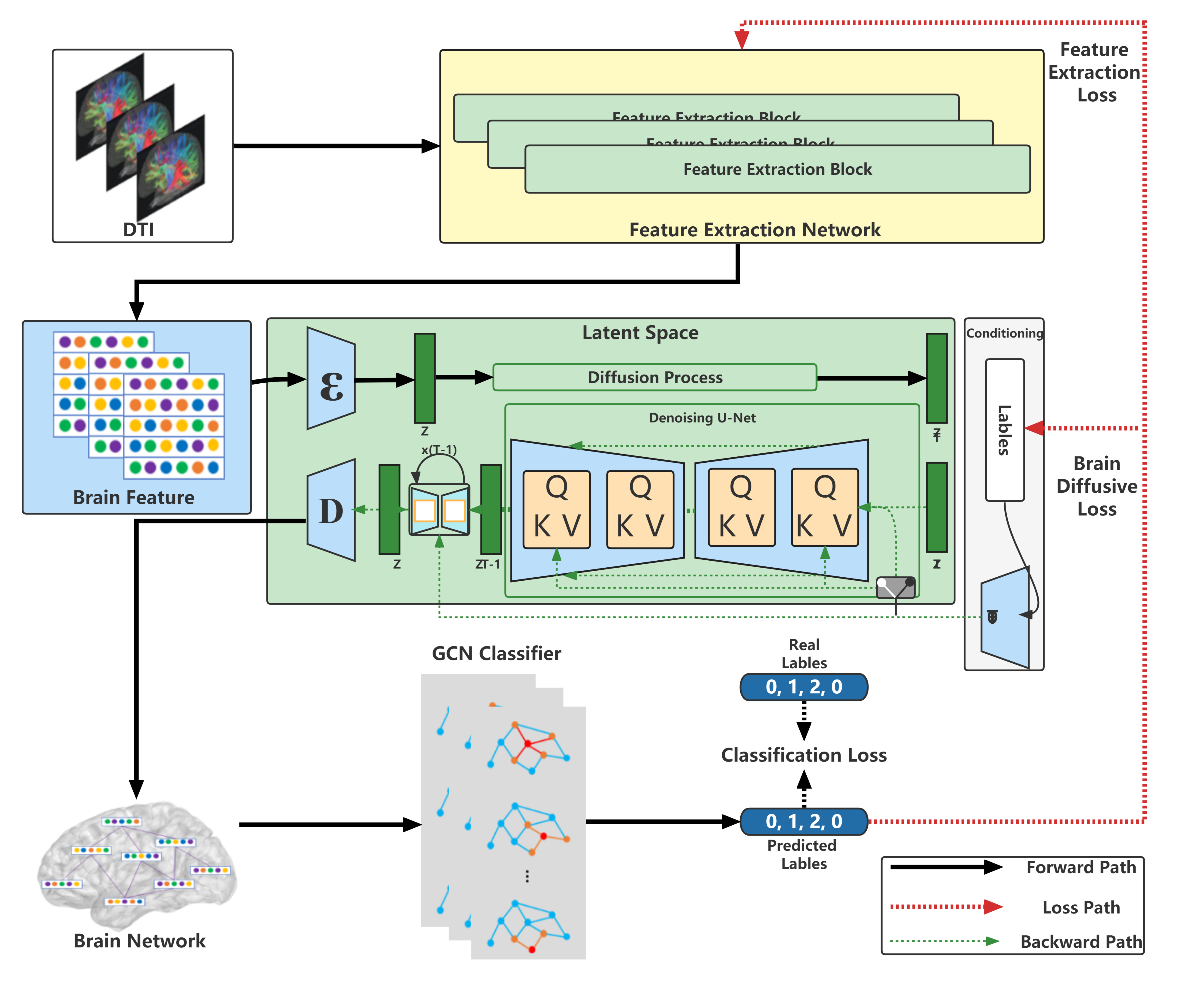}
    \end{minipage}
    \caption{
    The overall architecture of our Brain Diffuser. Feature extraction network is applied to DTI images to extract brain network topology features. After passing the features to the diffusion model, a complete brain network is synthesized. The GCN classifier will incorporate classification knowledge to improve feature extraction and diffusion model further.
    }
    \label{fig:model}
\end{figure*}

\section{Methodology}


\subsection{Feature Extraction Module}
To construct the brain network's topology, vertex and edge properties must be used. We propose the Feature Extraction Net (FENet) to extract these structural attributes from DTI.

Combining standard convolutional layer with depthwise separable convolutional blocks, with each block including a depth-wise convolutional layer and a pointwise convolutional layer, FENet retrieves features from DTI.

DTI is a single-channel, three-dimensional image with the shape $109\times91\times91$ after preprocessing. Each feature map serves as a structural feature vector of the relevant brain area, allowing us to derive the structural feature matrix $P\in R^{90\times80}$ of the brain region.

\subsection{Brain Diffuser}
As show in Figure~\ref{fig:model}, the proposed model is based on the latent diffusion model~\cite{rombach2022high}. It begins with an encoding module $\mathcal{E}$ and a decoding module $\mathcal{D}$. $\mathcal{E}$ encodes our structural feature matrix to latent space processed by the diffusion module. $\mathcal{D}$ decodes the latent diffusion model output into brain network $B\in R^{90\times90}$. These two modules concentrate on the crucial, semantic portion of the data and save a substantial amount of processing resources.

The primary component of this model is the capacity to construct the underlying U-Net~\cite{ronneberger2015u}, and to further concentrate the goal on the brain network topology using the reweighed bound as shown in Equation~\ref{eqn:ldm}:

\begin{equation}
L_{LDM}=E_{\mathcal{E}(x), \epsilon \sim \mathcal{N}(0, 1),  t} \Big[ \Vert \epsilon - \epsilon_\theta(z_{t},t) \Vert_{2}^{2}\Big]
\label{eqn:ldm}
\end{equation}

The network backbone $\epsilon_\theta(\circ, t)$ is a time-conditional U-Net. Since the forward process is fixed, $z_{t}$ can be efficiently obtained from $\mathcal{E}$ during training, and samples from $p(z$) can be decoded to image space with a single pass through $\mathcal{D}$.

To provide more adaptable results, we combine U-Net with the cross-attention technique~\cite{vaswani2017attention}.
We introduce a domain specific encoder $\tau_\theta$ hat projects input to an intermediate representation $\tau_\theta(y) \in R^{M\times d_\tau}$, which is then mapped to the intermediate layers of the U-Net via a cross-attention layer implementing $Attention(Q, K, V) = softmax(\frac{QK^T}{\sqrt{d}}) \cdot V$, with 

$$
    Q = W^{(i)}_Q \cdot  \varphi_i(z_t), \; K = W^{(i)}_K \cdot \tau_\theta(y),
  \; V = W^{(i)}_V \cdot \tau_\theta(y) 
$$

Here, $\varphi_i(z_t) \in R^{N \times d^i_\epsilon}$ denotes a (flattened) intermediate representation of the U-Net implementing $\epsilon_\theta$ and $W^{(i)}_V \in R^{d \times d^i_\epsilon}$, $W^{(i)}_Q \in R^{d \times d_\tau} $ \& $W^{(i)}_K \in R^{d \times d_\tau}$ are learnable projection matrices~\cite{jaegle2021perceiver}.

Based on input pairs, we then learn the conditional LDM via

\begin{equation}
L_{LDM} = E_{\mathcal{E}(x), y, \epsilon \sim \mathcal{N}(0, 1), t }\Big[ \Vert \epsilon - \epsilon_\theta(z_{t},t, \tau_\theta(y)) \Vert_{2}^{2}\Big] \, ,
\label{eq:learn}
\end{equation}
both $\tau_\theta$ and $\epsilon_\theta$ are jointly optimized via Eq.~\ref{eq:learn}. This conditioning mechanism is flexible as $\tau_\theta$ can be parameterized with domain-specific experts \textit{e.g.} the GCN classifier.

\subsection{GCN Classifier}
Our GCN classifier comprises of a GCN layer~\cite{kipf2017semi} and a fully connected layer. The GCN layer is used to fuse the structural characteristics of each vertex's surrounding vertices, while the fully connected layer classifies the whole graph based on the fused feature matrix. The classifier is used to categorize EMCI, LMCI, and NC, but its primary function is to guide the generation workflow utilizing classification information.

\subsection{Loss Function}

\subsubsection{Classification loss $L_C$}

Taking into account the structural properties of each vertex individually, we add an identity matrix $I$ to brain network $\hat{A}$. As demonstrated in Equation~\ref{eqn:gcn}, we utilize the output results of the classifier and the class labels to compute the cross-entropy loss as the loss of the classifier.

\begin{equation}
L_C = - \frac{1}{N} \sum_{i=1}^N{p(y_i|x_i) log[q(\hat{y}_i|x_i)]}
\label{eqn:gcn}
\end{equation}
Where $N$ is the number of input samples, $p\left(y_{i} \mid x_{i}\right)$ denotes the actual distribution of sample labels, and $q\left(\hat{y}_{i} \mid x_{i}\right)$ denotes the distribution of labels at the output of the classifier.

\subsubsection{Brain diffusive loss $L_B$}

As previously mentioned, the outcome of the diffusion model is contingent on the input and label-guided conditioning mechanism. FENet extracts the DTI feature as its input. And GCN classifier is applied to categorize brain networks and generate biomarkers that serve as labels. Consequently, Brain Diffuse Loss consists of Feature Extraction Loss $L_{FE}$, LDM Loss $L_{LDM}$, Classification Loss $L_C$. We use $L_1$ loss as $L_{FE}$ to retrieve disease related information according to the brain atlas. So our Brain Diffusive Loss $L_{B}$ can be described as Equation~\ref{eqn:diff}

\begin{equation}
    L_{B} = L_{FENet} + L_{LDM} + L_C
\label{eqn:diff}
\end{equation}

\section{Experiments}

\subsection{Dataset and Preprocessing}
Our preprocessing workflow began with employing the PANDA toolkit~\cite{cui2013panda} that used the AAL template. DTI data were initially transformed from DICOM format to NIFTI format, followed by applying head motion correction and cranial stripping. The 3D DTI voxel shape was resampled to $109\times91\times91$. Using the Anatomical Automatic Labeling (AAL) brain atlas, the whole brain was mapped into 90 regions of interest (ROIs), and the structural connectivity matrix was generated by calculating the number of DTI-derived fibers linking brain regions. Since the quantity of fiber tracts significantly varies across brain regions, we normalized the structural connectivity matrix between 0 and 1 based on the input's empirical distribution.

\begin{table}[ht]
\centering
\caption{ADNI dataset information in this study.}\label{tab1}
\begin{tabular}{P{0.23\linewidth}P{0.23\linewidth}P{0.23\linewidth}P{0.23\linewidth}}
\hline
\textbf{Group}   & \textbf{NC(87)} & \textbf{EMCI(74)} & \textbf{LMCI(31)}\\
\hline
Male/Female      & 42M/45F        & 69M/66F        & 35M/41F        \\
Age(mean$\pm$SD) & 74.1 $\pm$ 7.4 & 75.8 $\pm$ 6.4 & 75.9 $\pm$ 7.5 \\ 
\hline
\end{tabular}
\label{table:adni}
\end{table}

\begin{figure}[ht]
    \begin{minipage}[b]{1.0\linewidth}
        \begin{minipage}[b]{.32\linewidth}
            \centering
            \centerline{\includegraphics[width=\linewidth]{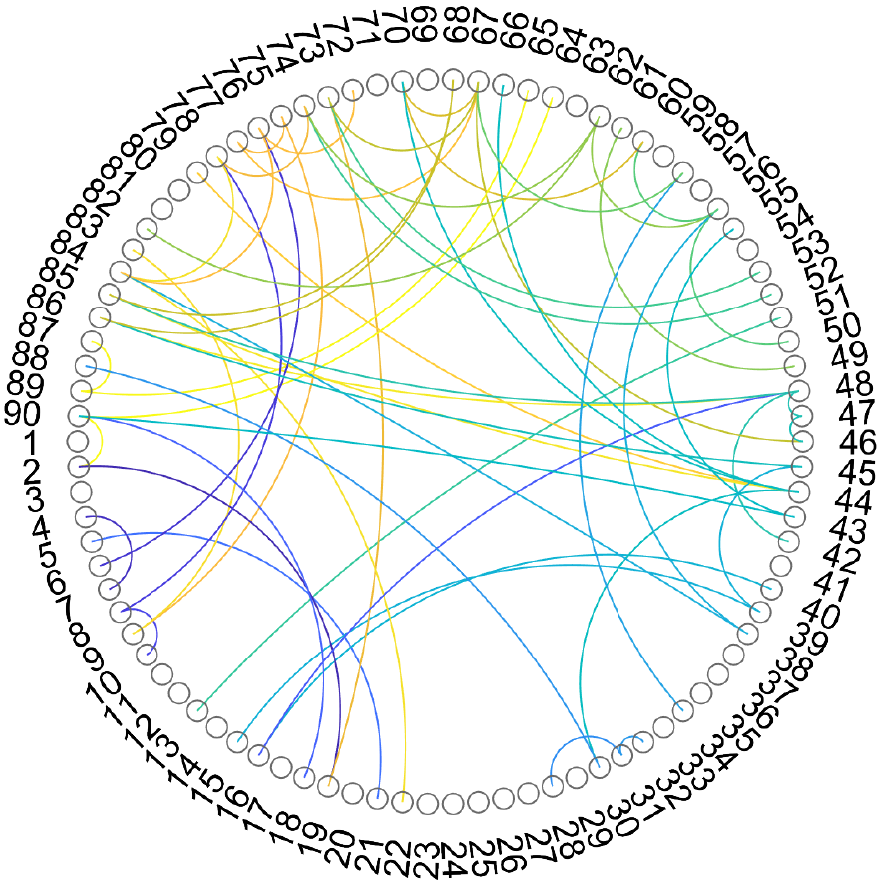}}
            \centerline{NC: $\hat{A}$ vs. $A$}\medskip
        \end{minipage}
        \hfill
        \begin{minipage}[b]{.32\linewidth}
            \centering
            \centerline{\includegraphics[width=\linewidth]{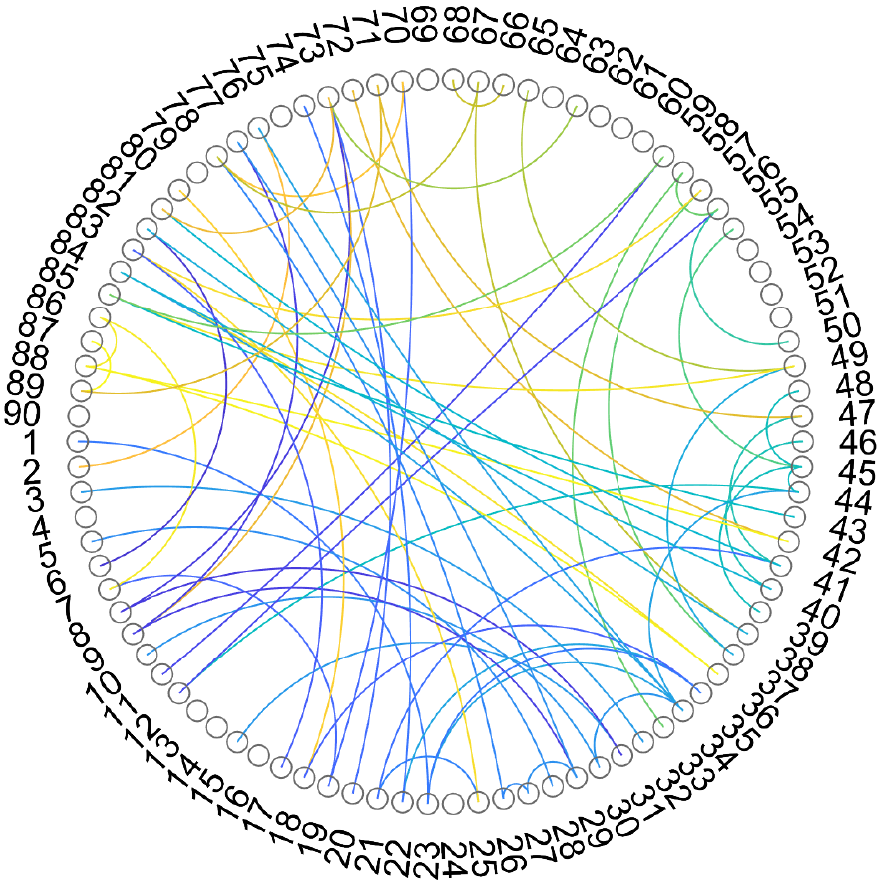}}
            \centerline{EMCI: $\hat{A}$ vs. $A$}\medskip
        \end{minipage}
        \hfill
        \begin{minipage}[b]{.32\linewidth}
            \centering
            \centerline{\includegraphics[width=\linewidth]{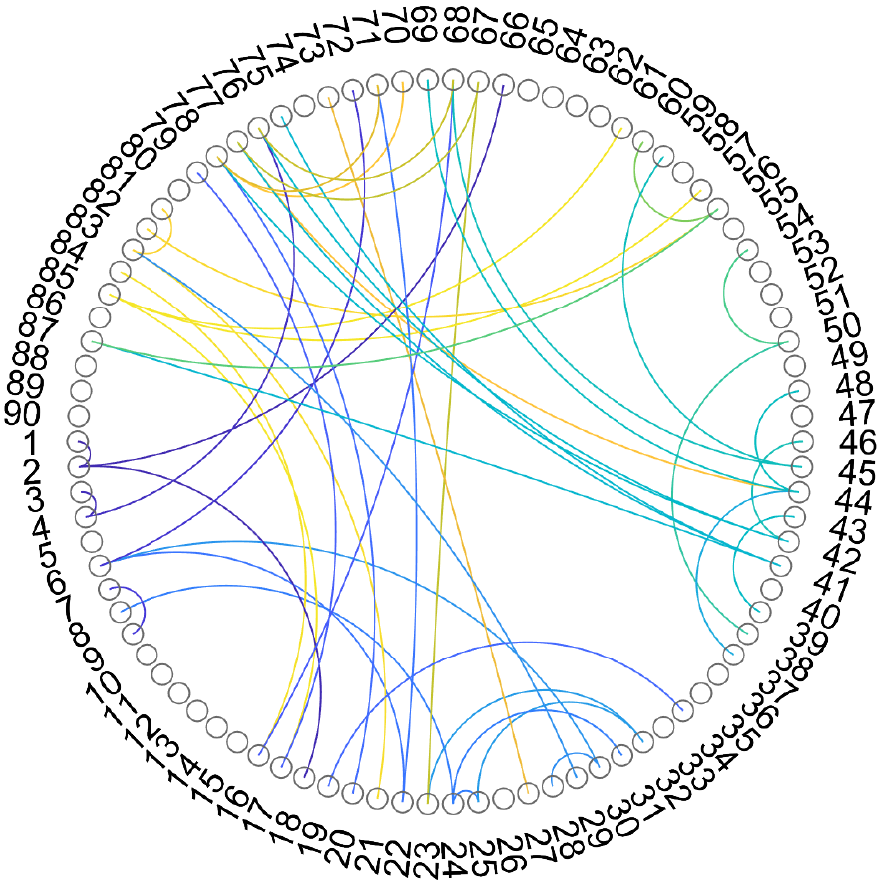}}
            \centerline{LMCI: $\hat{A}$ vs. $A$}\medskip
        \end{minipage}
    \end{minipage}
    \caption{Variations in structural brain connectivity at various developmental stages.}
    \label{fig:diff}
\end{figure}

\begin{figure}[ht]
    \begin{minipage}[b]{1.0\linewidth}
        \begin{minipage}[b]{.32\linewidth}
            \centering
            \centerline{\includegraphics[width=\linewidth]{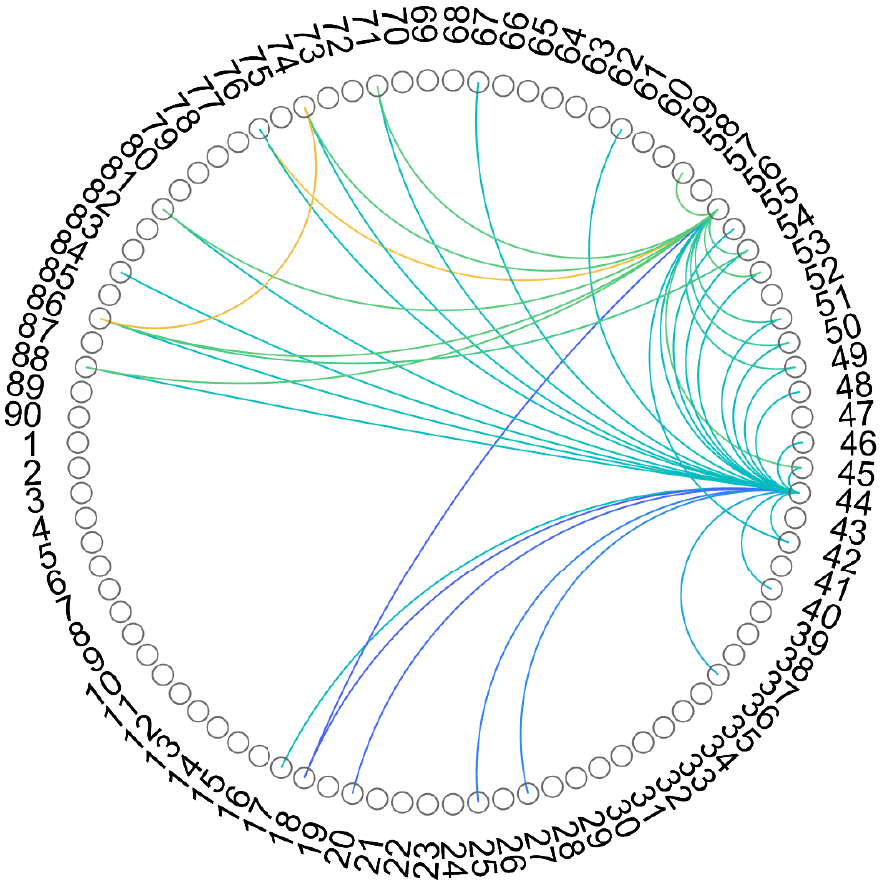}}
            \centerline{}\medskip
        \end{minipage}
        \hfill
        \begin{minipage}[b]{.32\linewidth}
            \centering
            \centerline{\includegraphics[width=\linewidth]{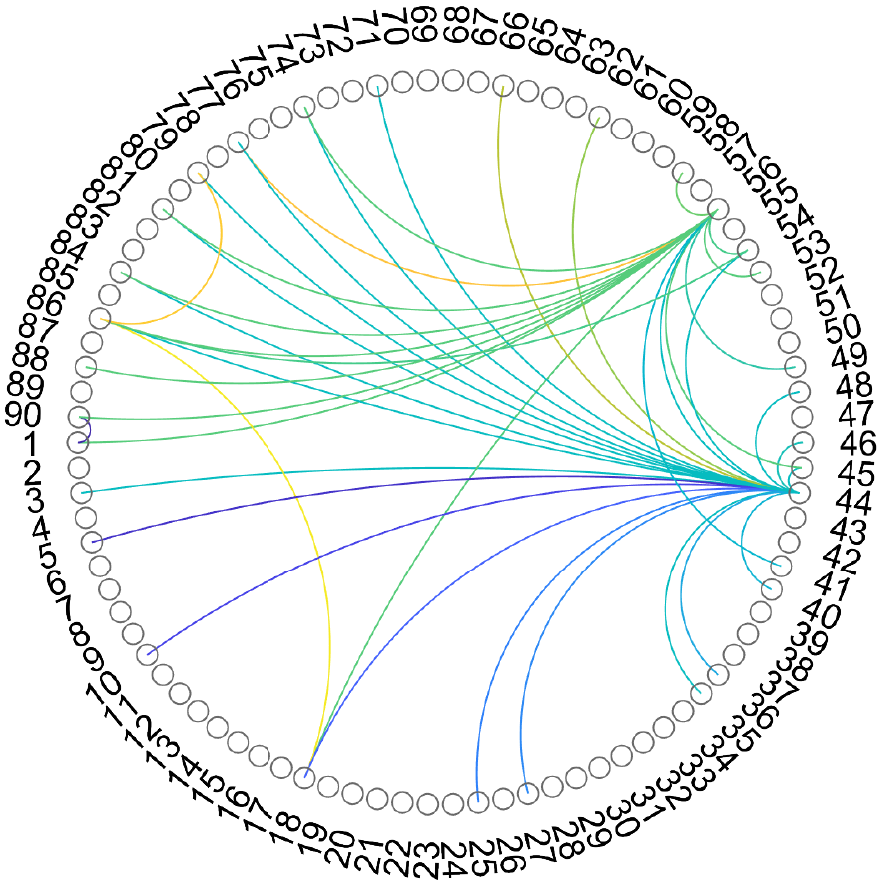}}
            \centerline{(a) Declined structural connections}\medskip
        \end{minipage}
        \hfill
        \begin{minipage}[b]{.32\linewidth}
            \centering
            \centerline{\includegraphics[width=\linewidth]{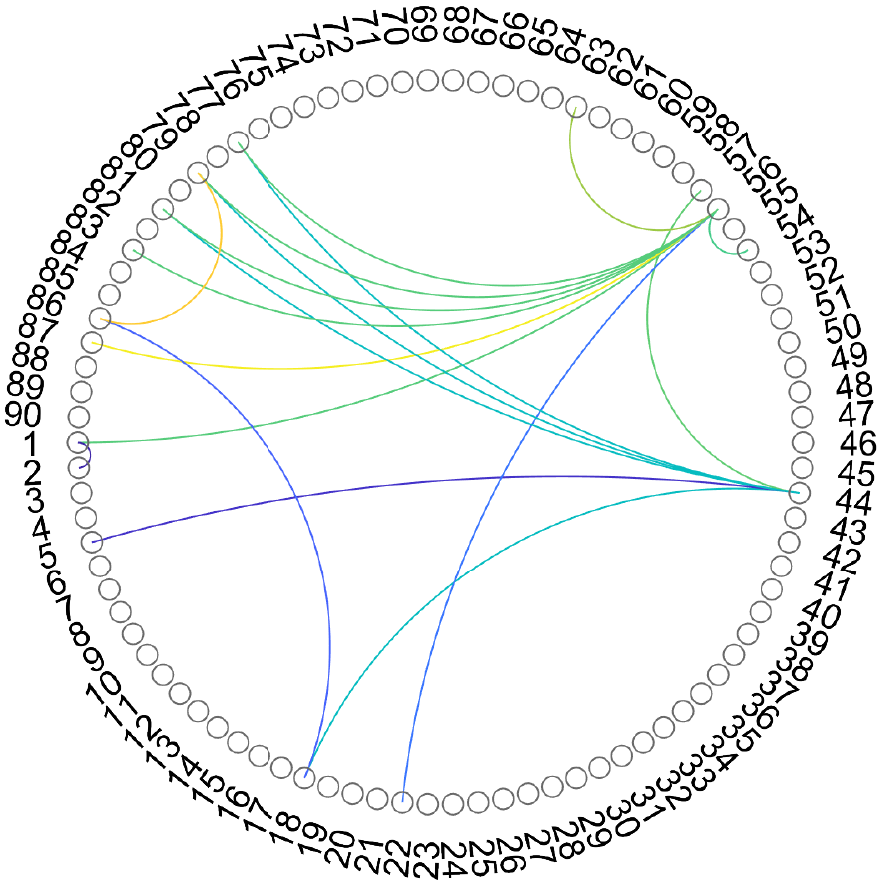}}
            \centerline{}\medskip
        \end{minipage}
    \end{minipage}
    \begin{minipage}[b]{1.0\linewidth}
        \begin{minipage}[b]{.32\linewidth}
            \centering
            \centerline{\includegraphics[width=\linewidth]{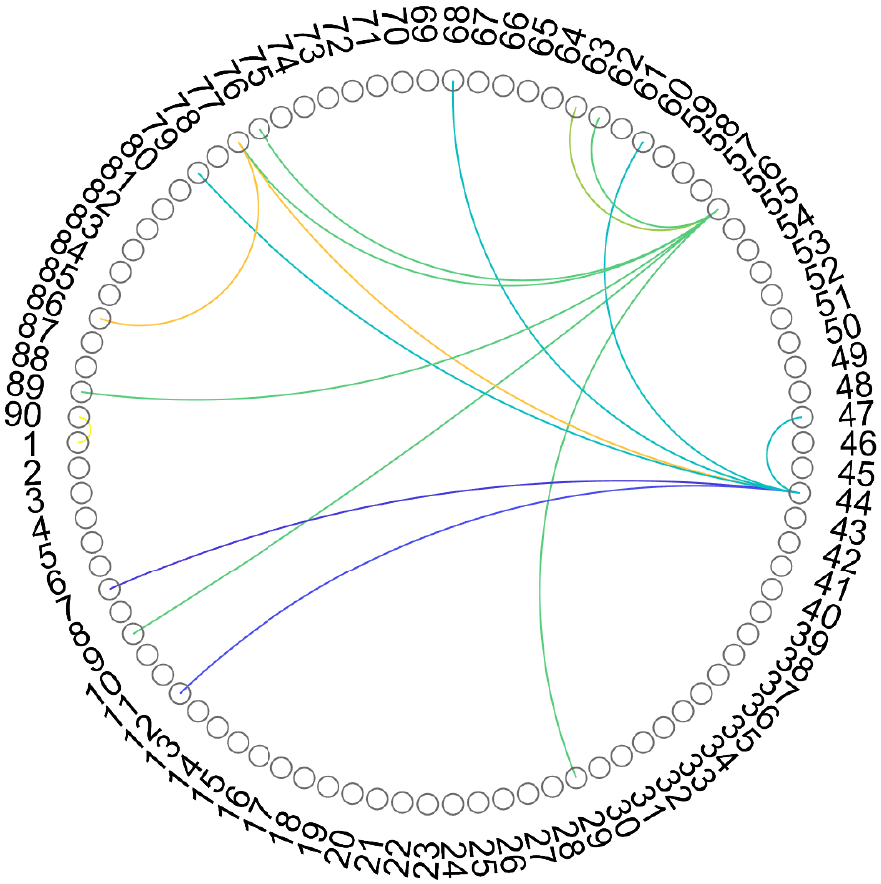}}
            \centerline{}\medskip
        \end{minipage}
        \hfill
        \begin{minipage}[b]{.32\linewidth}
            \centering
            \centerline{\includegraphics[width=\linewidth]{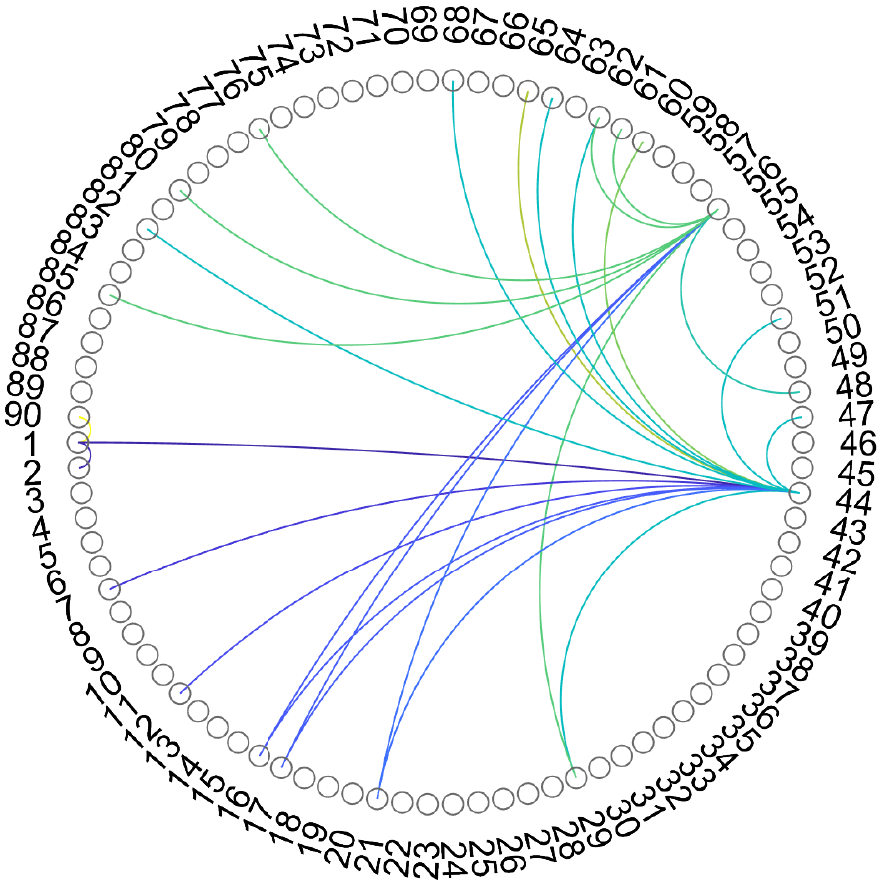}}
            \centerline{(b) Enhanced structural connections}\medskip
        \end{minipage}
        \hfill
        \begin{minipage}[b]{.32\linewidth}
            \centering
            \centerline{\includegraphics[width=\linewidth]{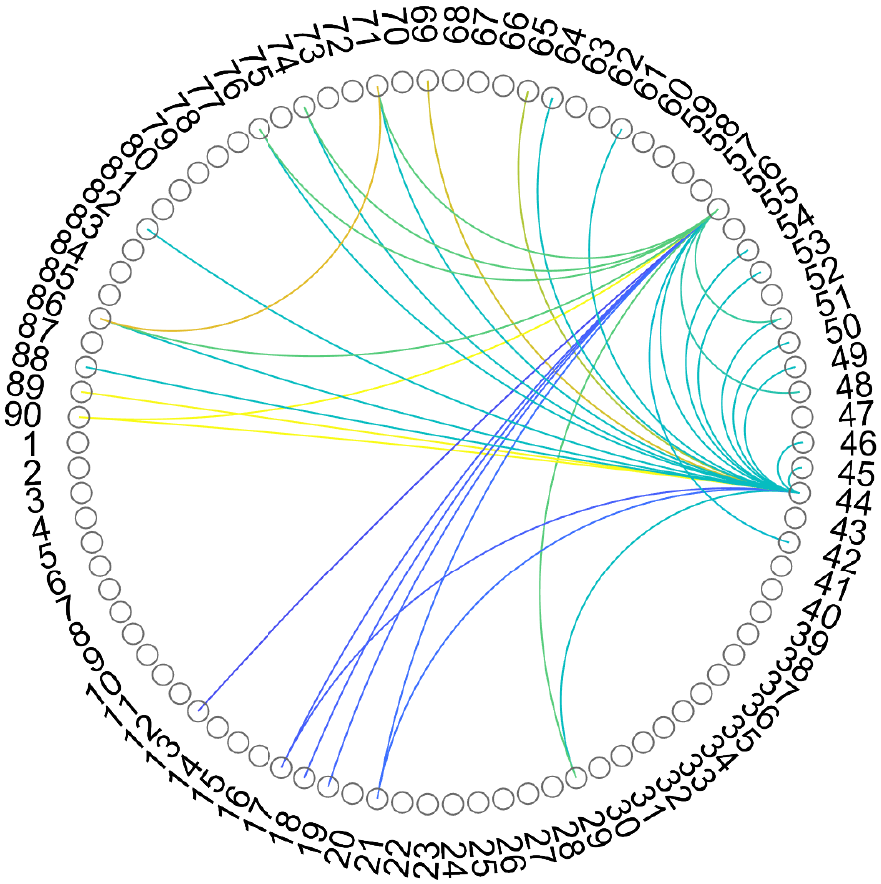}}
            \centerline{}\medskip
        \end{minipage}
    \end{minipage}
    \caption{Chord diagrams of connectivity change based on the empirical distribution of our synthesized connection matrix. From left to right: NC versus EMCI, NC versus LMCI, EMCI versus LMCI.}
    \label{fig:ours}
\end{figure}

\subsection{Experiment Configuration}
 We employed DTI, structural connectivity matrix A and each subject's matching label as model inputs. We divided 298 samples into groups of 268 training samples and 30 testing samples (9 for NC, 14 for EMCI, 7 for LMCI). The specific data information is presented in Table~\ref{table:adni}.

The performance of classification is defined by detection accuracy, sensitivity, specificity, and F1-score. We used PyTorch to implement our Brain Diffuser model. Our experiments were conducted using NVIDIA GeForce GTX 2080 Ti. The Adam optimizer was used with a batch size of 2, and we set the learning rate empirically to be $10^{-4}$.

\begin{table}[ht]
\centering
\caption{Prediction performance of the PANDA-generated brain networks and our reconstructed brain networks.}
\label{table:result}
\begin{tabular}{P{0.19\linewidth}P{0.19\linewidth}P{0.19\linewidth}P{0.19\linewidth}P{0.19\linewidth}}
\hline
      & Accuracy & Sensitivity & Specificity & F1-score \\ \hline
PANDA & 66.14\%  & 66.14\%     & 83.51\%     & 67.15\%  \\
Ours  & 87.83\%  & 87.83\%     & 92.66\%     & 87.83\%  \\ \hline
\end{tabular}%
\end{table}

\begin{table}[ht]
\centering
\caption{Ablation study on prediction performance of different classifiers.}
\label{table:ablation}
\begin{tabular}{P{0.19\linewidth}P{0.19\linewidth}P{0.19\linewidth}P{0.19\linewidth}P{0.19\linewidth}}
\hline
          & Accuracy & Sensitivity & Specificity & F1-score \\ \hline
GAT~\cite{velickovic2018graph}       & 25.66\%  & 25.66\%     & 63.15\%     & 21.44\%  \\
GIN~\cite{xu2018how}       & 75.93\%  & 75.93\%     & 87.72\%     & 73.31\%  \\
Graphormer~\cite{ying2021transformers}   & 80.69\%  & 80.69\%     & 88.31\%     & 77.38\%  \\
GraphSAGE~\cite{Hamilton2017InductiveRL} & 74.87\%  & 74.87\%     & 86.72\%     & 72.89\%  \\ \hline
Ours      & 87.83\%  & 87.83\%     & 92.66\%     & 87.83\%  \\ \hline
\end{tabular}%
\end{table}

\begin{figure}[ht]
    \begin{minipage}[b]{1.0\linewidth}
        \begin{minipage}[b]{.49\linewidth}
            \centering
            \centerline{\includegraphics[width=\linewidth]{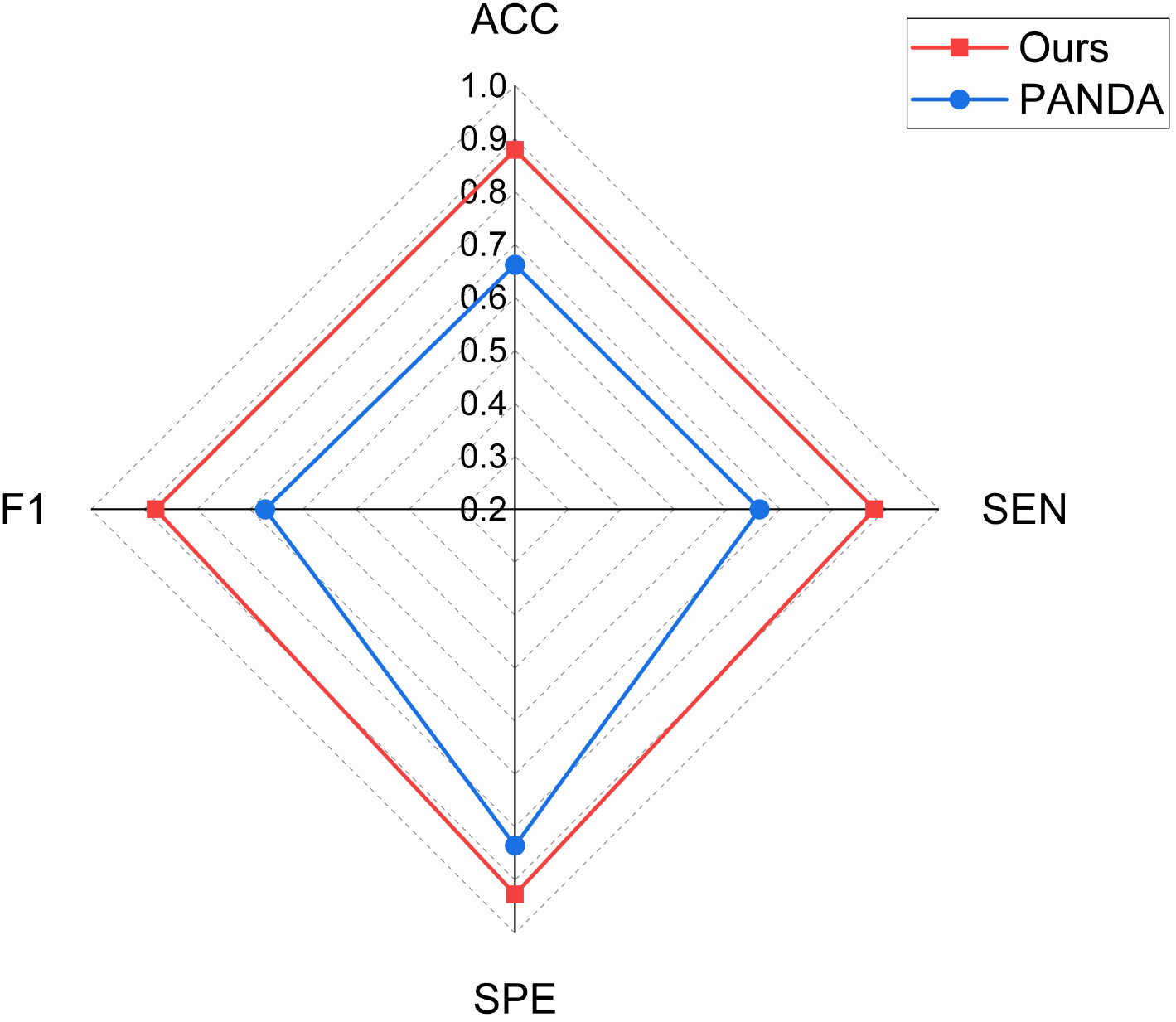}}
        \end{minipage}
        \hfill
        \begin{minipage}[b]{.49\linewidth}
            \centering
            \centerline{\includegraphics[width=\linewidth]{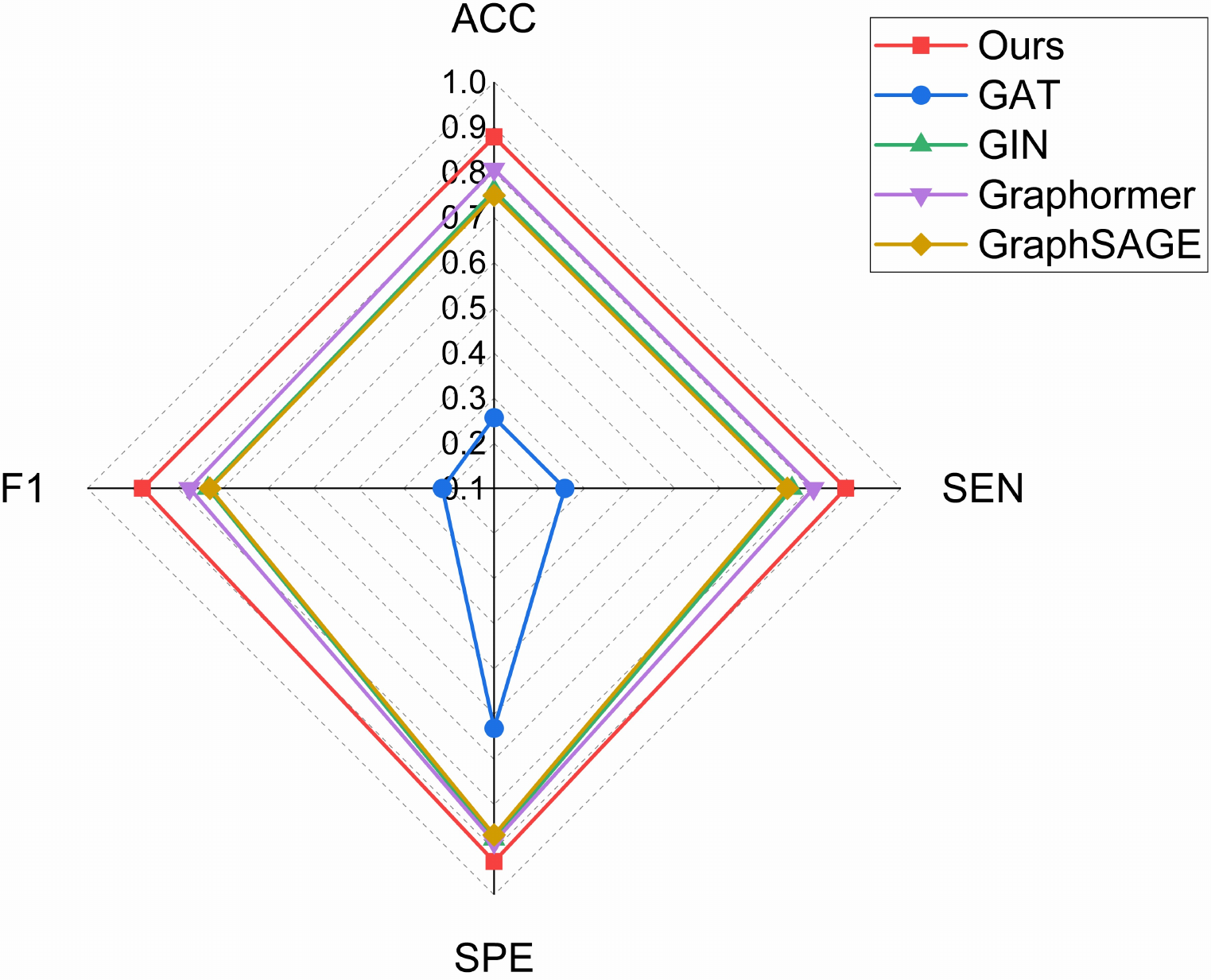}}
        \end{minipage}
    \end{minipage}
    \caption{Radar charts of classification performance.}
    \label{fig:radar}
\end{figure}

\subsection{Results and Discussion}

\subsubsection{Classification performance}

The results presented in Table~\ref{table:result} and Figure~\ref{fig:radar} demonstrate that our reconstructed results outperform those generated by the PANDA toolkit. This suggests that our reconstructed pipeline has more advantages for classifiers detecting the potential biomarkers of MCI. Our generator's performance is significantly superior to that of PANDA, which has great potential for future research in AD detection and intervention. The ablation study in Table~\ref{table:ablation} and Figure~\ref{fig:radar} demonstrate that our classifier outperforms others, suggesting that our classifier can effectively utilize a greater number of disease-related features or prospective disease-related biomarkers.

\subsubsection{Brain structural network connectivity analysis}

To determine whether the structural connectivity matrix $\hat{A}$ generated by our proposed model differed significantly from the PANDA outcome $A$, we conducted a paired-samples T-test at a threshold value of 0.05 between $\hat{A}$ and $A$ following Kong \emph{et al.}'s setting~\cite{kong2022adversarial}. The chord diagram in Figure~\ref{fig:diff} indicates connections with significant changes. The structural connectivity matrix $\hat{A}$ derived by our model has altered connectivity between numerous brain regions compared to the static software generation.

To establish whether there was a significant difference between the structural connectivity matrix $\hat{A}$ generated by our proposed model and the PANDA result $A$, we performed a paired-samples T-test at a threshold value of 0.05 between $\hat{A}$ and $A$ using Kong \emph{et al.}'s setting~\cite{kong2022adversarial}. The connections that exhibited substantial changes are illustrated in the chord diagram in Figure~\ref{fig:diff}. Our model generated structural connectivity matrix $\hat{A}$ demonstrated significant variations in connectivity between numerous brain regions compared to the static software generation.

Figure~\ref{fig:ours} compares the changes in brain connectivity in different stages. The alteration in connectivity between brain regions was more pronounced in LMCI stage than in NC. Similarly, there was a significant decline in connectivity between brain regions in LMCI patients compared to EMCI patients. These alterations and tendencies reveal a sequential progression toward AD pathology in NC subjects: a progressive decrease in structural brain network connectivity.

\section{Conclusion}

This paper presents Brain Diffuser, a novel approach for directly generating structural brain networks from DTI. Brain Diffuser provides an entire pipeline for generating structural brain networks from DTI that is free of the constraints inherent in existing software toolkits. Our method enables us to study structural brain network alterations in MCI patients. Using Brain Diffuser, we discovered that the structural connectivity of the subjects' brains progressively decreased from NC to EMCI to LMCI, which is in accordance with previous neuroscience research. Future studies will locate which brain regions of AD patients exhibit the most significant changes in structural connectivity.

%
%
%
\bibliographystyle{splncs04}
\bibliography{ref}
\end{document}